# MedSensor: Medication Adherence Monitoring Using Neural Networks on Smartwatch Accelerometer Sensor Data

Chrisogonas Odhiambo
Computer Science & Engineering
University of South Carolina
Columbia, SC, USA
odhiambo@email.sc.edu

Pamela Wright
College of Nursing
University of South Carolina
Columbia, SC, USA
shealypw@email.sc.edu

Cindy Corbett
College of Nursing
University of South Carolina
Columbia, SC, USA
corbett@sc.edu

Homayoun Valafar
Computer Science & Engineering
University of South Carolina
Columbia, SC, USA
homayoun@cse.sc.edu

**Abstract** – *Poor medication adherence presents serious economic and health problems including compromised treatment effectiveness, medical complications, and loss of billions of dollars in wasted medicine or procedures. Though various interventions have been proposed to address this problem, there is an urgent need to leverage light, smart, and minimally obtrusive technology such as smartwatches to develop user tools to improve medication use and adherence. In this study, we conducted several experiments on medication-taking activities, developed a smartwatch android application to collect the accelerometer hand gesture data from the smartwatch, and conveyed the data collected to a central cloud database. We developed neural networks, then trained the networks on the sensor data to recognize medication and non-medication gestures. With the proposed machine learning algorithm approach, this study was able to achieve average accuracy scores of 97% on the protocol-guided gesture data, and 95% on natural gesture data.*

## I. INTRODUCTION

Poor adherence to prescription medication is a major problem with a myriad health and economic implications. It can lead to compromised treatment effectiveness, medical complications, and even death especially when strict adherence to medication dosage and frequency is required. It can also lead to loss of billions of dollars in unnecessary health care expenses due to wasted medicine or further health complications arising from poor medication adherence [1]–[3]. Studies show that 33-69% of all medication-related hospital admissions in the United States (US) are caused by poor medication adherence, which translates to an annual cost of approximately $100 billion [4], [5].

Annually in the US, non-adherence can account for up to 50% of treatment failures, approximately 125,000 deaths, and up to 25% of hospitalizations [6]. Typically, adherence rates of 80% or more are needed for optimal therapeutic efficacy. However, it is estimated that adherence to chronic medications is around 50% [6].

The two main causes of poor medication adherence are stress and the complexity of medication procedure or steps [7]. Both physical and emotional stress on a patient may result in depression, anger, denial of illness, or fear of medication. The complexity includes factors such as the dosage, frequency, duration, cost, and refill policy, which can demotivate the patient. While stress may be difficult to control by external factors or tools, the complexity burden may be reduced by technology.

With these economic and health implications, it is imperative to provide tools and means to enable medication adherence. The question then becomes: What are some of the readily available and affordable tools that can leverage modern technology to support adherence to medication? The purpose of this study was to explore the use of smartwatch sensors in monitoring human hand motions to detect medication-taking, with the aim to help people adhere to their prescriptions, hence minimizing the negative effects of poor medication. This, in conjunction, with other messaging technologies such as Amazon Alexa, or simple SMS notifications, can provide useful medication reminders.

## II. BACKGROUND
### A. Wearables in Human Activity Detection

Various studies demonstrate how smartwatches have been used to monitor and detect human motions, such as the case of smoking detection [8]–[11], or fall-detection [12]–[15]. Independent reports [16]–[21] also confirm the usability of smartwatches and other smart wearables in the study of complex human motion behaviors such as eating habits, physical activities, and foot motion [16]–[25]. Considering the rich array of sensors, cost, accessibility, and ease of use, smartwatches have emerged as a compelling platform to unobtrusively study human activities. Uses of Smartwatches include step-counters [22], sleep monitoring [23], diet monitoring [18] as well as general fitness tracking [24]. Smartwatches have demonstrated [25], [26] high accuracy for detecting smoking gestures [8], [25], [26]. Smoking gestures were detected with 95% accuracy in the laboratory environment[27] and 90% accuracy in-situ [8]. Studies to identify smoking via gestures has also been demonstrated to be more accurate when compared to self-reporting (90% versus 78%)[8], [28].

### B. Wearables in Detection of Medication Adherence

Monitoring medication-taking can be broadly categorized as direct or indirect. The former involves observation of a person e taking medicines or drug-testing in a laboratory [7]. The latter involves self-reporting, pill counting, medication refill tracking, and electronic tracking using smart wearables, cameras or pill caps with medication event monitoring systems [1], [4], [29]. Direct methods are most accurate, but generally more obtrusive, time-consuming, and expensive. Indirect methods are relatively inexpensive, efficient, and less obtrusive tools for monitoring and reporting medication-taking. This study focuses on indirect approaches of medication monitoring.

Smart wearable devices have been utilized in indirect observation of medication adherence in numerous ways including: (1) self-reporting facilitated by mobile devices[30], (2) sensors worn around neck such as the SenseCam[31] was originally envisaged for use within the domain of Human Digital Memory to create a personal lifelog or visual recording of the wearer's life, which can be helpful as an aid to human memory, (3) multi-axis inertial sensors worn on wrists[32], [33], and (4) the use of commodity smartwatches [34]. In summary, all the approaches have collectively demonstrated the potential of smart sensors to promote medication adherence, but leave potential for improvement in performance, cost, convenience, and usability.

## III. DATA COLLECTION AND METHOD

The overall approach to our investigation included data collection from human subjects (n=31) followed by developing and testing the performance of Artificial Neural Networks to identify medication-taking events. The following sections provide the details for each step of our studies.

### A. Data Collection Protocol

The data collection was performed using Wear OS compatible smartwatches worn on each participant's right wrist. A custom software package named MedSensor was installed on the smartwatches to facilitate data collection, annotation, storage, and transmission. The acquired data consisted of time stamp and x, y, and z components of the accelerometer data sampled at 25Hz intervals. In addition, information regarding the start and end of each medication-taking session was provided by the user and recorded to assist with data annotation. Participants marked the beginning and end of each medication-taking activity by pressing the corresponding button (shown in Figure 1) on the MedSensor app. A total of 31 participants were included in the data collection activities. Each participant was directed to record 10 protocol-guided medication-taking activities per day for 5 days, followed by 5 days of recording 10 medication-taking activities per day using their natural medication-taking gestures. In total 1300 protocol-guided and 1300 natural medication-taking gestures were collected.

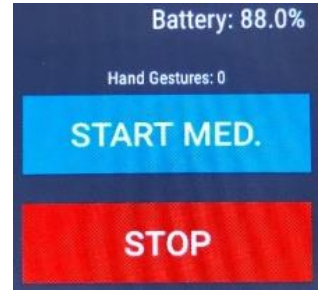

*Figure 1. MedSensor display assisting the users with the annotation of the start and end of each gesture.*

### B. Medication Taking Activity

Our data collection consisted of two broad categories of protocol-guided and natural medication-taking gestures. The first phase of our study (presented here) focused on the recognition of the protocol-guided medication-taking activity as a proof of concept. The protocol-guided medication-taking activity is defined as the five explicit consecutive steps shown in Table 1. The natural gesture medication-taking activity is purely defined by the participant and likely consists of many permutations of the sub-activities shown in Table 1 and performed by any combination of left or right hands. It is important to note that medication-taking gestures cannot be performed by a single hand and it must involve the use of both hands.

*Table 1. Protocol-guided medication-taking activity.*

| Step | Description | Activity |
|---|---|---|
| 1 | Unscrew the medicine bottle cap with your right hand while holding the bottle by your left hand. | 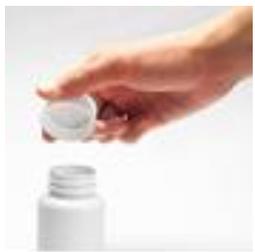 |
| 2 | Tip the medicine bottle with left hand to dispense pill(s) onto your right hand. | 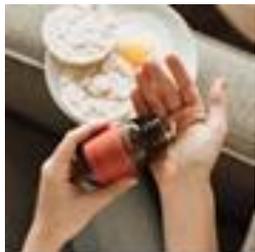 |

| 3 | Place/toss pill to mouth using the right hand. | 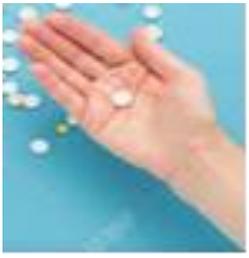 |
|---|---|---|
| 4 | Pick up beverage/drink with the right hand, bring to mouth and drink to swallow "pill". | 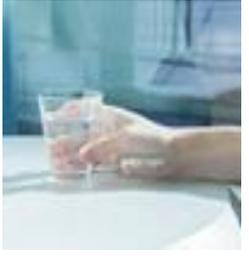 |
| 5 | Set glass down, hold the medicine bottle in left hand, and put its cap back using right hand. | 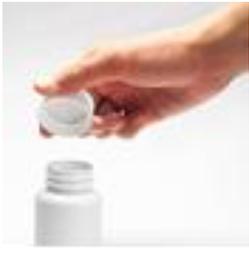 |

### C.     Data Consolidation and Transfer

Upon completion of medication-taking recording sessions, each participant submitted the medication-taking gesture data collected by the watch to their paired smartphone via Bluetooth. Further, the participant submitted the data from the phone to the cloud storage via an internet connection and the MedSensor phone application. The phone provided the bridge between the watch and cloud because the watches did not provide for direct file upload to the cloud. Besides this role, the phones were not necessary. Other than data transfer to cloud through the phone, it was also possible to access watch data directly via data cables or Wi-Fi. However, such a method would not be practical for some participants. To establish a homogenous protocol, the participants were directed to use the MedSensor interface to collect and submit data to a centralized cloud storage. This was a better locationally transparent process that also preserved the integrity of data from capture to dispatch. The data from the watch is a zip of two csv files: actual sensor data and annotation points that identify the medication gestures and the non-medication gestures. It is important to note that the data collected includes non-medication gestures. These are equally important in the network training since they ultimately help the models discern what is a valid and what is not a valid medication gesture.

### D.     Data Annotation Process

Careful and proper data annotation is one of the most critical, time-consuming, and challenging aspects of utilizing supervised learning. In our study, we used the self-reported start and end of each medication-taking event to easily expose a small section of a person's recorded medication-taking gestures. Figure 2 shows an example of one activity of interest embedded within a larger recording session. The self-reported start and end annotations are rough approximations and require further scrutiny by a trained supervisor. Aside from human error in marking the beginning and end of an activity by a user, the recorded data can include unrelated activities such as the gesture that is required to mark the START/END on the smartwatch. Therefore, we further refined the start and end of the medication-taking activity while confirming the existence of one. This process produces a more reliable and accurate set of training and testing data. A typical medication gesture is shown in Figure 2. Although in our current investigation we use the entire medication taking gesture as one complete signal, in principle, it is possible to subdivide and analyze the signal as a temporal sequence of sub-gestures as denoted by segments A-D in Figure 2. These segments correspond to open-bottle dispense-medicine (A), Hand-to-mouth pill-to-mouth hand-off-mouth (B), pick-up-water drink-water lower-cup-to-table close-bottle (C), and post-medication (D), respectively. Figure 3 is a superimposition of three separate medication-taking events collected from the same participant.

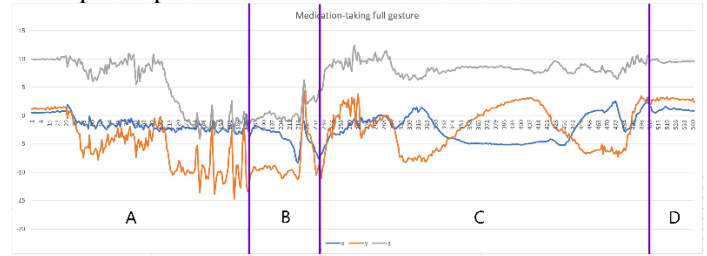

*Figure 2: Visualization of a full medication gesture from one participant.*

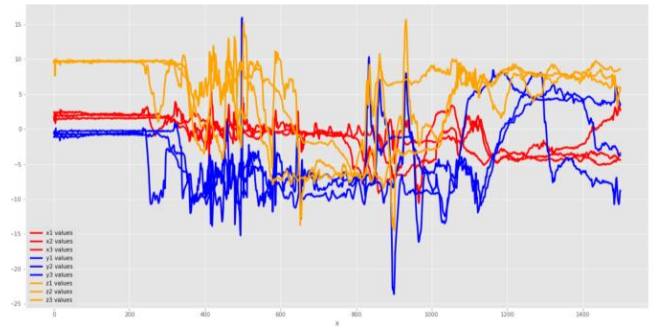

*Figure 3: Visualization of the superimposition of three medication gestures from one participant.*

### E. ANN Training and Testing Methodology

We explored three related aims in our investigations. These aims are as follows and gradually span the gamut of the intended applications of our software:

Exp #1. Explore the capabilities of the network when trained and tested with protocol-guided data from all participants. The test data was a split in-sample dataset from the protocol-guided dataset.

Exp #2. Explore the capabilities of the network when trained with the protocol-guided data from n-1 participants to be tested with the n-th participant. Both the train and test datasets were protocol-guided gestures except the test dataset that was an out-of-sample.

Exp #3. A preliminary exploration of training a network using all gestures (both the protocol-guided and natural) from n-1 participants and tested on the n-th participant. The training dataset also included the protocol guided dataset of the n-th participant.

In all these experiments, the explored ANNs were presented with an entire gesture. Therefore, the input size of ANNs consisted of the longest medication-taking gesture across all the training and testing sets. This consisted of 1500 consecutive accelerometer data points (approximately 20 seconds) that required an input size of 4500 neurons to accommodate the x, y, and z components of the accelerometer. The output layer consisted of a single neuron reporting the presence or absence of a medication-taking event. We employed a parsimonious strategy in determining the size and number of hidden layers. During each phase of our experimentation, an array of hidden neurons was explored to yield the optimal performance. In all the experiments, a single hidden layer sufficed (general architecture shown in Figure 4), and we therefore did not explore deep architectures. Generally, the number of hidden neurons started at 10 and was incremented in steps of 10 to as many as 100 hidden neurons. In each application, the optimal architecture was then selected to be carried for testing purposes.

In experiment #1 the training and testing data sets were randomly selected from all the data in the ratio of 80:20 respectively. Each bootstrapping exercise (Exp #2, and #3) was repeated n-times by excluding each participant in each round. We used Keras/TensorFlow platform for all our ANN simulations. A loss function of accuracy defined in Eq 1 was used to assess the performance of the trained ANN. In this equation the terms TP, TN, FP, and FN correspond to true positive, true negative, false positive, and false negative, respectively.

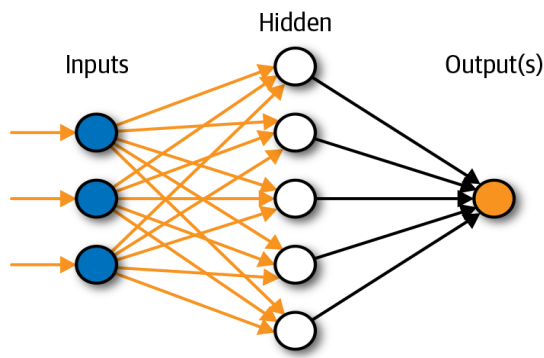

*Figure 4: Artificial Neural Network high-level architecture*

$$Accuracy = \frac{TP+TN}{TP+FP+TN+FN} \qquad \text{Eq (1)}$$

## IV. RESULTS AND DISCUSSION

### A. Results of Experiment #1

The first experiment was the most fundamental test that could be conducted. The results of this experiment are shown in Table 2. Each experiment was repeated three times to assess whether results varied due to randomization of the training/testing datasets. In general, the results did not vary noticeably and therefore, we report the results of one single instance of training/testing. Based on the results shown in this table, when developing a single ANN to detect a medication-taking event across the entire sample, a hidden layer size of 90 neurons is the optimal architecture. Although not a large reduction, the performance of ANN slightly decreases with a larger or smaller hidden layer size.

Based on these results, it is possible that a single ANN (with optimal architecture of 4500, 90, 1) can detect a medication-taking event with as high as 97.8% accuracy if the 31 participants in our study provided a comprehensive representation of all medication-taking events across the entire population. Skeptical of this conclusion, we embarked on evaluation of ANNs in detecting medication-taking events for new participants in the next experiment.

*Table 2. Results of a two-layer ANN as a function of hidden neurons for Experiment #1.*

| Hidden Neurons | Accuracy Training | Accuracy Testing |
|---|---|---|
| 100 | 98.90% | 95.99% |
| 90 | 98.95% | 97.77% |
| 80 | 98.58% | 95.99% |
| 80 | 98.96% | 95.99% |
| 60 | 98.25% | 97.18% |
| 50 | 98.93% | 97.48% |

| 40 | 98.41% | 96.59% |
|---|---|---|
| 30 | 99.08% | 97.18% |
| 20 | 98.24% | 97.48% |
| 10 | 99.06% | 96.59% |
| **Max** | 99.08% | 97.77% |
| **Min** | 98.24% | 95.99% |
| **Average** | 98.74% | 96.82% |

### B. Results of Experiment #2

To better address the generalizability and practical applications of the presented detection mechanism, we tested the ability of ANN to identify the protocol-guided medication-taking event for a new participant. This approach allows the immediate use of the developed application on any new user without the need to retrain the network. Therefore, ANNs were trained using bootstrapping to train a network on n-1 participants and testing with the n-th out-sample/participant dataset using the protocol-guided medication-taking events. This experiment was repeated 10 times for each of 31 participants by altering the hidden neurons from 10 to 100 in increments of 10; in total, 310 ANNs were examined. Table 3 and Table 4 show summary of results for the first 10 of the 31 participants, for both training and testing accuracy scores, respectively, for the best and worst architectures. The **Hidden Neurons** column shows the number of hidden neurons in the configurations that produced the highest and lowest train or test accuracies. The averages in both tables refer to the average for the 31 participants. As an example, in Table 3, for Participant 1 (row 1), the best training performance was achieved by the configuration 4500-**20**-1 while best test performance (according to the corresponding Table 4) was achieved by the configuration 4500-**40**-1. The lowest corresponding train/test scores from the two tables were recorded by the configurations 4500-**30**-1 and 4500-**80**-1, respectively.

Several conclusions can be derived from the results shown in the two tables. First, detection of medication-taking events from new participants is possible with accuracies varying from 98% (participant 1) to 100% (participants 2 and 7) with an average of 99.7% across all participants. However, the optimal performance corresponds to a different number of hidden neurons for each participant. This anecdotal observation agrees with the general expectation of human behavior where some people may exhibit a more complex behavioral signature while others exhibit a simpler behavioral signature. The complex signatures require a more capable ANN, which translates to a greater number of hidden neurons. The results shown in Table 5 are the average performance of each ANN configuration across the entire sample. Based on these results, an ANN with 60 or 100 hidden neurons exhibits an average performance of 96.8% across all participants and therefore, while not optimally configured for any one participant, they perform consistently well across our entire cohort. It further shows that no one model was the best fit for all participants. This was perhaps due to the fact that each participant's hand motions have some degree of uniqueness or signature as illustrated by Figure 2 and Figure 3. The latter shows that gestures from the same participant vary. However, there is clearly an emergent motion pattern in all the gestures.

*Table 3: Training Accuracy Results of ANN training using a bootstrap approach after experimenting with 10 different hidden layer sizes for each excluded participant. The average is across the 31 participants.*

| Partici-pant | Accuracy Scores | | Hidden Neurons Count for | |
|---|---|---|---|---|
| | **Highest** | **Lowest** | **Highest Accuracy** | **Lowest Accuracy** |
| 1 | 97.48% | 96.30% | 20 | 30 |
| 2 | 98.10% | 94.44% | 60 | 50 |
| 3 | 98.10% | 90.51% | 90 | 10 |
| 4 | 97.51% | 78.54% | 20 | 10 |
| 5 | 96.45% | 94.29% | 90 | 30 |
| 6 | 97.54% | 95.54% | 20 | 70 |
| 7 | 97.54% | 92.78% | 10 | 90 |
| 8 | 96.78% | 95.55% | 30 | 90 |
| 9 | 98.77% | 79.29% | 30 | 10 |
| 10 | 97.70% | 97.09% | 40 | 10 |
| **AVG** | **97.49%** | **93.60%** | | |

*Table 4: Testing Accuracy Results of ANN training using a bootstrap approach after experimenting with 10 different hidden layer sizes for each excluded participant. The average is across the 31 participants.*

| Partici-pant | Accuracy Scores | | Hidden Neurons Count for | |
|---|---|---|---|---|
| | **Highest** | **Lowest** | **Highest Accuracy** | **Lowest Accuracy** |
| 1 | 97.99% | 31.66% | 40 | 80 |
| 2 | 100.00% | 84.82% | 10 | 90 |
| 3 | 100.00% | 79.44% | 50 | 10 |
| 4 | 100.00% | 100.00% | 10 | 10 |
| 5 | 100.00% | 100.00% | 10 | 10 |
| 6 | 100.00% | 100.00% | 10 | 10 |
| 7 | 100.00% | 100.00% | 10 | 10 |
| 8 | 100.00% | 84.21% | 50 | 10 |
| 9 | 98.18% | 69.09% | 70 | 10 |
| 10 | 94.34% | 77.36% | 60 | 10 |
| **AVG** | **99.11%** | **86.79%** | | |

Table 5: The average performance of ANNs for each architecture.

| Hidden Neurons | Average Training Accuracy (%) | Average Testing Accuracy (%) |
|---|---|---|
| 100 | 96.76% | 93.25% |
| 90 | 96.40% | 96.49% |
| 80 | 96.54% | 95.42% |
| 70 | 96.79% | 97.25% |
| 60 | 96.77% | 97.12% |
| 50 | 96.52% | 97.79% |
| 40 | 96.64% | 96.48% |
| 30 | 96.42% | 96.59% |
| 20 | 96.82% | 96.90% |
| 10 | 94.93% | 94.50% |

## C. Results of Experiment #3

This experiment was conducted with an out-sample natural gesture dataset as follows: The training dataset comprised of all protocol-guided data of n-participants plus (n-1) natural datasets. The $n^{th}$ natural gesture dataset was used as the test set. Table 6 shows the highest, lowest and average accuracy scores as well as the number of observations used in the training and testing procedure for the first cohort of participants who completed the data collection process successfully (n=10).

For experiment #3, using the data from participants' natural gestures, we also tested all the 10 configurations as was the case with the protocol-guided data.

Table 6

| Participant | Highest (%) | | Lowest (%) | | Averages (%) | | Dataset Sizes | |
|---|---|---|---|---|---|---|---|---|
| | Train | Test | Train | Test | Train | Test | Train | Test |
| User1 | 97.7 | 98.3 | 96.5 | 94.7 | 97.3 | 97.7 | 5228 | 57 |
| User2 | 97.8 | 98.2 | 96.9 | 69.6 | 97.3 | 89.1 | 5230 | 56 |
| User3 | 97.3 | 98.2 | 77.5 | 46.3 | 94.7 | 91.3 | 5234 | 54 |
| User4 | 97.1 | 100 | 95.0 | 69.6 | 96.4 | 94.8 | 5250 | 46 |
| User5 | 97.5 | 100 | 95.6 | 84.8 | 96.7 | 96.5 | 5250 | 46 |
| User6 | 97.4 | 100 | 96.3 | 77.8 | 97.0 | 85.8 | 5172 | 45 |
| User7 | 97.2 | 100 | 96.1 | 100 | 96.9 | 100 | 5254 | 44 |
| User8 | 97.5 | 100 | 96.0 | 84.0 | 96.8 | 98.4 | 5292 | 25 |
| User9 | 97.5 | 100 | 95.7 | 95.8 | 96.8 | 96.3 | 5294 | 24 |
| User10 | 97.6 | 100 | 96.0 | 82.4 | 96.7 | 93.5 | 5308 | 17 |
| **AVG** | **97.5** | **99.5** | **94.2** | **80.5** | **96.7** | **94.3** | | |

Note that all tests were conducted on the trained models with out-sample natural gesture datasets; the sample was excluded from the model training.

## V. CONCLUSIONS

Three experiments were conducted to identify participants' medication-taking gestures. We demonstrated that we could leverage smartwatches non-obtrusively to harness the power of technology to consistently identify medication-taking gestures. Neural networks predicted medication-taking gestures that were greater than 97% accurate for protocol-guided data and 95% accurate for natural medication-taking gestures. Importantly, we were able to establish that every person has some uniqueness in medication-taking hand-motions. We trained different fitting models to suit each of these unique characteristics. The ability to accurately identify medication-taking gestures has the potential to improve medication adherence monitoring and translate to better population health outcomes and reduced health care costs. Combining medication reminders through SMS notifications or the use of conversational agents such as Amazon Echo may be particularly effective to improving medication adherence rates.

## VI. FUTURE WORK

In our future work, we intend to evaluate distinct parts of the medication-taking gestures as well as consider the full gesture. Correct recognition of parts of the whole may better distinguish medication-taking gestures from other similar gestures such as drinking in the absences of medication-taking. This may be better achieved using Long Short-Term Memory (LSTM) recurrent neural networks which are architecturally suited for order and sequence prediction problems. Finally, this study was done with a single smartwatch worn on the wrist of the right hand. This meant that majority of the recorded and analyzed hand motions were based on the right-hand motions. In a few cases, we observed gestures that showed less pronounced motions. It is possible that in such cases, the participant wore the watch on the wrist that was not executing the actual medication motions for the natural gesture, Therefore, for future studies and for a more comprehensive analysis and characterization of medication gestures, it will be useful to consider concurrent data collection using two smart watches on both hands.

## VII. ACKNOWLEDGMENTS

This work was supported by NIH grant number P20 RR-016461 to Dr. Homayoun Valafar, the SmartState **A**dvancing **C**hronic Care **O**utcomes through **R**esearch and i**N**novation (ACORN) Center, College of Nursing, University of South Carolina (Dr. Cynthia Corbett,

Director), and Ms. Wright's NIH grant number 1F31NR019206-01A1.